\documentclass[letterpaper, 10 pt, conference]{ieeeconf}  

\IEEEoverridecommandlockouts                              

\usepackage{graphicx} 
\usepackage{amssymb}  

\usepackage{multirow}
\usepackage{bm}
\usepackage{mathtools} 
\usepackage{float}
\usepackage{booktabs}
\usepackage{stfloats}
\usepackage[hidelinks]{hyperref}

\title{\LARGE \bf
Structured Multitask Gaussian Processes for Probabilistic Full-Body Human Motion Prediction}

\author{{Jinger Chong}$^{1}$, {Xiaotong Zhang}$^{1}$, {Kamal Youcef-Toumi}$^{1}$
\thanks{*This research was made possible by the support and partnership of King Abdulaziz City for Science and Technology (KACST) through the Center for Complex Engineering Systems at Massachusetts Institute of Technology (MIT) and KACST.}
\thanks{$^{1}${Department of Mechanical Engineering}, {Massachusetts Institute of Technology}, {Cambridge, MA 02139, USA}.
        {\tt\small {jinger@mit.edu}, {kevxt@mit.edu}, {youcef@mit.edu}}}%
}

\begin{document}

\maketitle
\thispagestyle{empty}
\pagestyle{empty}

\begin{abstract}
Robots that share space with people need human motion predictions with reliable uncertainty, not just accurate means. This paper presents a structured multitask variational Gaussian process (GP) for probabilistic full-body human motion prediction. The model integrates one-shot prediction of the full horizon, multitask outputs that couple future time steps, a factorization into one sparse variational GP per joint--dimension pair of the rotation representation, and a continuous 6D rotation representation with forward kinematics that preserves bone lengths. Its predictive distribution is available in closed form: a Gaussian with full temporal covariance for each rotation coordinate, which downstream planners operating in rotation coordinates can use without sampling, density estimation, or forward kinematics. On Human3.6M, the model achieves lower kernel-density-estimated negative log-likelihood (KDE NLL) than Motron and DLow at every prediction step, 22--52~nats below Motron. Empirical coverage is conservative at the 50\% and 80\% levels and close to nominal at 95\% over a 2~s horizon. The probabilistic model uses 0.24~M parameters, about seven times fewer than Motron, and a separately trained 0.35~M-parameter deterministic variant has a mean angle error 3--22\% higher than Motron's. Ablations confirm the benefit of the 6D representation and show that, at a fixed training budget, factorized joint outputs outperform coupled right-arm alternatives in NLL and CRPS while using less memory. These results show that compact, structured GPs provide analytic, temporally structured uncertainty with conservative empirical coverage for robotic systems.
\end{abstract}

\section{INTRODUCTION}

Human motion prediction plays a crucial role in applications such as autonomous driving, sports analysis, and human--robot collaboration (HRC). As robots increasingly operate in shared environments with humans, accurately anticipating human motion has become essential for safety. In tasks such as collaborative assembly or assistive care, predicting human movements enables robots to plan ahead, make informed decisions, and avoid potential collisions. The stochastic and multimodal nature of human motion presents a fundamental challenge, making uncertainty quantification as critical as prediction accuracy itself. Calibrated uncertainty lets a planner size safety margins to the actual risk: overconfident predictions invite collisions, while overly conservative ones make the robot needlessly slow~\cite{tcplanner, safetyaware}. Low point-prediction error alone does not provide this information. Despite substantial progress, many current methods prioritize point accuracy over the quality of the predicted distribution, or rely on models with millions to tens of millions of parameters.

Probabilistic approaches address this challenge by predicting human motion as distributions rather than deterministic outcomes. Recent state-of-the-art (SOTA) probabilistic methods leverage deep learning architectures, including graph-based networks~\cite{motron}, transformers~\cite{spatio}, and diffusion models~\cite{diffusionprob}, to capture complex, multimodal motion patterns. These methods provide flexible learned distribution families and demonstrate strong predictive performance. We examine whether a structured Gaussian process (GP) can retain competitive distribution quality while providing analytic, temporally structured predictive distributions with a compact parameterization.

GPs provide inherent uncertainty estimates, but GP-based motion prediction has been limited to low-dimensional or partial-body data. For example, the approach in~\cite{aadi} predicts the motion of three joints per arm, using a separate autoregressive GP for each joint angle, Jacobian-based mappings to task space, and Monte Carlo rejection sampling to enforce kinematic and workspace constraints. While physically consistent, this design relies on sequential rollouts and sampling whose cost grows with the number of joints and prediction steps, and it has been demonstrated only on arm trajectories and relatively small datasets.

Building on this work, we develop a structured GP framework for full-body human motion on a large-scale dataset. Its novelty lies not in any single component but in their integration: one-shot prediction of the full horizon, multitask~\cite{multitask} outputs that capture temporal correlations across future steps, sparse variational inference~\cite{inducing} with one GP per joint--dimension pair for scalability, and a continuous 6D rotation representation that suits the GP's smoothness assumptions and avoids the discontinuities and ambiguities of Euler angles and quaternions. The result is a compact model whose principal advantage is not mean prediction alone, but an analytically available, temporally structured predictive distribution with empirically validated coverage that downstream robotic systems can use directly.

In this work, we make the following contributions: (i) to our knowledge, the first GP framework for full-body human motion prediction on a large-scale dataset, Human3.6M (H3.6M)~\cite{h36m}, enabled by this integration; (ii) a closed-form Gaussian predictive distribution with full temporal covariance for each rotation coordinate, together with a precise statement of when consumers can avoid sampling, density estimation, and forward kinematics; (iii) an evaluation showing that the 0.24~M-parameter probabilistic model achieves lower kernel density estimate negative log-likelihood (KDE NLL) than Motron~\cite{motron} and DLow~\cite{dlow} with conservative-to-near-nominal coverage, that a 0.35~M-parameter deterministic variant has a mean angle error 3--22\% higher than Motron's, and the measured inference cost of both; (iv) ablations supporting the 6D representation, the hyperparameter choices, and the factorized design; and (v) a public preprocessing pipeline that reconstructs the legacy exponential map archive of H3.6M data~\cite{martinez2017human} and includes verification and 3D visualization tools, enabling reproducible research.

Collectively, these results position a compact, structured GP as a source of analytic, temporally structured uncertainty with empirically validated coverage that achieves lower KDE NLL than Motron and DLow on H3.6M while approaching Motron's point accuracy at longer horizons.

\section{RELATED WORK}

\subsection{Probabilistic Human Motion Prediction} 

Human motion prediction methods can be broadly categorized as deterministic, stochastic, or probabilistic. Deterministic approaches predict a single future trajectory and often fail to capture the inherent variability of human motion over longer horizons~\cite{scenediverse}. To address this, stochastic methods generate multiple potential futures~\cite{spatio, dlow, transfusion, scenediverse}. However, these methods often prioritize diversity over plausibility, limiting their applicability in robotics tasks that require both accuracy and reliability~\cite{motron, diffusionprob}.

Probabilistic approaches provide a principled way to explicitly model uncertainty through probability distributions over future trajectories. This enables both estimation of multiple potential outcomes and the integration of confidence measures into downstream tasks such as motion planning, improving safety and execution time in HRC~\cite{tcplanner, safetyaware, bayesian}. Existing probabilistic strategies include Bayesian networks~\cite{bayesian}, graph-based networks~\cite{motron}, diffusion models~\cite{diffusionprob}, invertible networks~\cite{ma2025uncertainty}, and GPs~\cite{gpdm}. Although these approaches support expressive predictive distributions, methods that provide only samples require an additional density-estimation step for likelihood evaluation. We adopt a structured GP formulation with a closed-form Gaussian predictive distribution, providing direct access to density and temporal covariance in rotation coordinates over the full horizon.

\subsection{Gaussian Processes for Dynamical Systems}

GPs are widely used to model nonlinear dynamical systems, but conventional GPs scale poorly with high-dimensional inputs and outputs, which has limited them to partial-body human motion~\cite{aadi}.

To address scalability, several approaches learn lower-dimensional latent representations, as in GP-LVMs~\cite{gpdm, damianou2011variational}. While effective for reducing dimensionality, latent representations express uncertainty in a learned space rather than directly in joint coordinates. In contrast, our framework maintains the original joint representation and instead factorizes the GP at the joint--dimension level, enabling tractable computation for full-body motion. Conventional GP approaches also often treat outputs as independent, overlooking dependencies that have been shown to improve performance across applications, including leg and upper-body motion capture forecasting~\cite{vardependent}. We address this by incorporating multitask GP extensions~\cite{multitask} to capture temporal correlations across the prediction horizon.

Another limitation of standard GP models is their cubic complexity with respect to the number of observations, which restricts their use to relatively small datasets. To overcome this, we employ sparse variational approximations with inducing points~\cite{inducing}, enabling efficient training on a large-scale human motion dataset. Together, these design choices allow scalable and temporally coherent full-body motion prediction with explicit kernel assumptions.

\subsection{Pose Representations for Human Motion Prediction}

Human poses can be represented in multiple ways, each with trade-offs. Position-based representations, such as 3D joint coordinates~\cite{dlow, guo2023back, chen2023humanmac}, are straightforward to optimize but may violate bone-length constraints, producing physically implausible poses. The nonlinear manifold defined by fixed bone lengths can also make Euclidean kernels poorly reflect pose differences, affecting accuracy and uncertainty calibration.

Rotation-based representations in $\mathrm{SO}(3)$, including Euler angles~\cite{aadi}, exponential maps~\cite{ma2025uncertainty, martinez2017human}, and quaternions~\cite{motron, pavllo2020modeling}, preserve kinematic consistency but introduce challenges for regression. Euler angles are discontinuous, quaternions lie on a manifold, and Euclidean distances in these spaces may poorly reflect actual rotational differences. These limitations have motivated our use of smooth, continuous embeddings for rotation regression in human motion prediction.

\subsection{Model Efficiency in Human Motion Prediction}

Human motion prediction models vary widely in size. Deterministic models range from about 0.1~M parameters for siMLPe~\cite{guo2023back} to several million for HisRepItself~\cite{mao2020history}, stochastic models typically use 7--30~M~\cite{dlow, chen2023humanmac, motiondiff}, and probabilistic models use roughly 0.3--2~M~\cite{motron, ma2025uncertainty}. Parameter count is only a rough proxy for computational cost, and GPs additionally require kernel evaluations and matrix factorizations, so parameter counts and measured inference cost are reported separately.

\section{PROBLEM FORMULATION}

The goal of this work is to predict probability distributions over human joint poses $\bm{s}_i \in \mathbb{R}^{J \times D}$, where $J$ denotes the number of joints defining the human skeleton and $D$ specifies the dimensionality of the chosen pose representation for each joint. At time step $t$, given past observations $\bm{x}_{t} = \{\bm{s}_{t-H+1}, \dots, \bm{s}_{t}\}$, with $H$ as the observation window, we aim to estimate the distributions of future poses $\bm{y}_{t} = \{\bm{s}_{t+1}, \dots, \bm{s}_{t+F}\}$ over a prediction horizon of $F$ time steps. Formally, this can be expressed as $p(\bm{y}_t \mid \bm{x}_t)$. Depending on the experimental setup and evaluation metric, the input and output pose representations, along with their dimensionalities, may vary.

\section{METHODOLOGY}

\subsection{Architecture Overview}

Human motion prediction is often formulated in two ways: autoregressive rollout~\cite{ma2025uncertainty, aadi} or one-shot forecasting~\cite{motron}. 

In the autoregressive setting, future poses are generated sequentially
\begin{equation}
p(\bm{y}_t \mid \bm{x}_t) = \prod_{f=1}^{F} p(\bm{s}_{t+f} \mid \bm{x}_t, \bm{s}_{t+1:t+f-1})
\end{equation}
where each predicted step depends on all previously predicted steps. While this explicitly models temporal dependencies, uncertainty accumulates recursively, often producing over-dispersed predictions at longer horizons. 

In contrast, one-shot forecasting predicts all future steps jointly
\begin{equation}
p(\bm{y}_t \mid \bm{x}_t) = p(\bm{s}_{t+1:t+F} \mid \bm{x}_t),
\end{equation}
yielding a single joint distribution over the prediction horizon. We adopt this approach to avoid compounding uncertainty and directly capture correlations across future steps.

\begin{figure}[thpb]
  \centering
  \setlength{\fboxrule}{0pt}
  \fbox{\includegraphics[width=.975\linewidth]{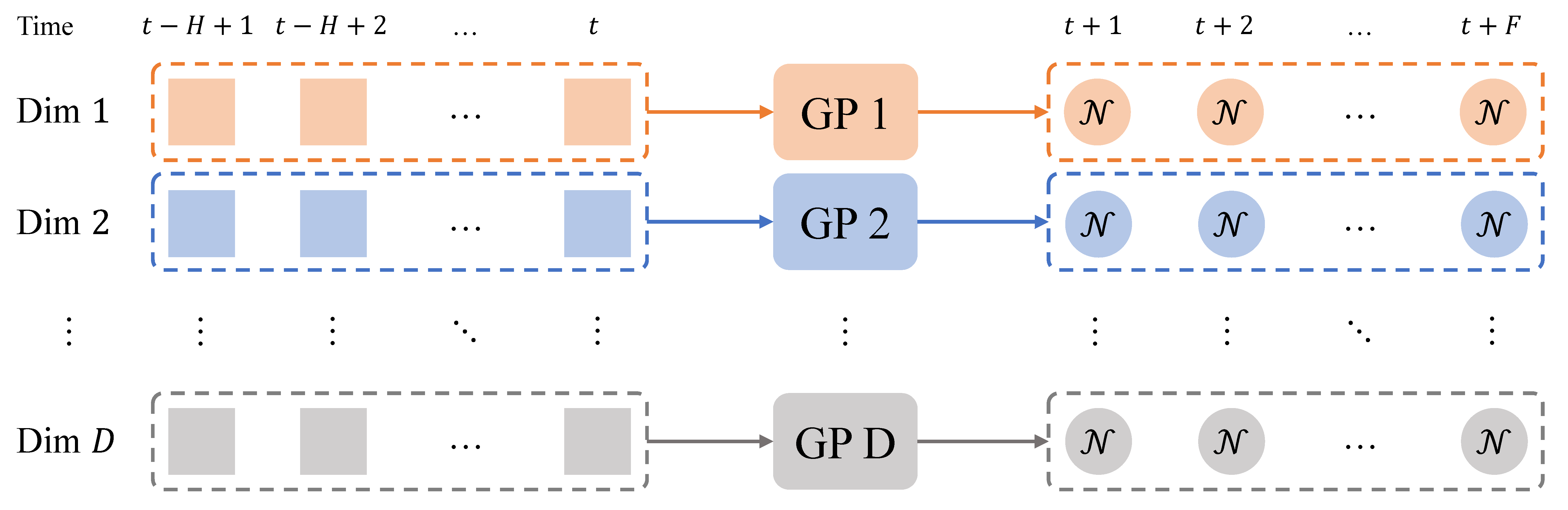}}
  \caption{Model architecture illustrated for a single joint with $D$ dimensions. Each joint--dimension pair is modeled by a GP that maps $H$ past time steps to $F$ future time steps and produces a Gaussian predictive distribution at each future time step. This structure is replicated for all joints, resulting in 96 parallel GPs after preprocessing. The factorization keeps full-body inference tractable while each GP retains a full temporal covariance over the prediction horizon.}
  \label{fig:arch}
\end{figure}

We factorize the one-shot prediction across joint--dimension pairs, modeling each pair with a separate GP as shown in Fig.~\ref{fig:arch}.
\begin{equation}
f_{j,d}: \mathbb{R}^{H} \rightarrow \mathbb{R}^{F}
\end{equation}

Each GP maps $H$ past observations to a Gaussian distribution over $F$ future steps, capturing temporal correlations within that horizon. After removing zero-variance dimensions, 96 independent GPs remain, without explicit cross-joint or cross-dimension covariance. The joint coupling ablation in Section~\ref{sec:coupling} evaluates this factorization against within-joint and full-chain coupling.

\textbf{Computational Motivation for Factorization.} A full-body GP maps $\mathbb{R}^{H \times J \times D} \rightarrow \mathbb{R}^{F \times J \times D}$, giving 6000 output dimensions for $J=20$, $D=6$, and $F=50$. Even after pruning to $P=96$ joint--dimension pairs, a dense joint predictive covariance has $4800\times4800$ entries (92~MB in fp32 per sequence, about 189~GB for a batch of 2048 sequences), which factorization avoids.

\subsection{GP Formulation}

Building on the one-shot prediction framework, each GP is multitask, jointly predicting multiple output dimensions. To capture temporal correlations, the $F$ outputs of each GP are coupled via a linear model of coregionalization (LMC) with $L$ latent functions.

To scale to large datasets, we adopt sparse variational approximation~\cite{inducing}, introducing $M$ inducing variables. This reduces the computational complexity from $O(N^3)$ to $O(N M^2)$ per GP, where $N$ is the number of training samples.

Each GP produces an analytical Gaussian distribution over its $F$ future steps,
\begin{equation}
\bm{y}_{j,d} \sim \mathcal{N}(\bm{\mu}_{j,d}, \bm{\Sigma}_{j,d}),
\end{equation}
as illustrated in Fig.~\ref{fig:arch}, where the LMC induces a full temporal covariance $\bm{\Sigma}_{j,d}\in\mathbb{R}^{F\times F}$. The full-body prediction is thus Gaussian over all normalized rotation coordinates and time steps, with a block-diagonal covariance that models temporal but not cross-joint or cross-dimension dependencies. Affine unnormalization preserves this form, whereas Gram--Schmidt orthonormalization and forward kinematics (FK) do not. Consumers operating on the 6D coordinates can therefore use the means and covariances directly, without sampling, density estimation, or repeated FK. Position-space consumers, including the position-space metrics in Section~V, rely on Monte Carlo samples. Calibration is assessed empirically in Section~\ref{sec:prob}.

Finally, input similarity is encoded using a Matérn 3/2 kernel with an additive linear term
\begin{equation}
k(\bm{x},\bm{x}') = \sigma^2 \left(1 + \frac{\sqrt{3}r}{\ell}\right)\exp\left(-\frac{\sqrt{3}r}{\ell}\right) + \sigma^2_\text{lin} \bm{x}^\top \bm{x}'
\label{eq:matern_kernel}
\end{equation}
where $r = |\bm{x}-\bm{x}'|$, $\ell$ is the lengthscale, $\sigma^2$ the Matérn variance, and $\sigma^2_\text{lin}$ scales the linear contribution. The Matérn term captures local smoothness, while the linear term accounts for long-term drift in normalized input space. These kernel parameters make the model's inductive bias explicit.

\subsection{Pose Representation}

We encode each rotation by stacking the first two columns of its matrix into a 6D vector~\cite{zhou2019continuity}. Differentiable Gram--Schmidt orthonormalization maps predicted vectors back to valid rotations. This smooth, continuous embedding makes Euclidean distances approximate local rotational differences, supporting the GP kernel's smoothness assumptions. FK with fixed skeleton offsets then reconstructs joint positions and preserves bone lengths by construction.

\section{EXPERIMENTS}

\subsection{Dataset}
\label{sec:data}

We evaluated our approach on H3.6M~\cite{h36m}, which contains Vicon motion capture recordings of 11 subjects performing 15 everyday activities at 50~Hz. It provides parent-relative intrinsic ZXY Euler angles in degrees for 25 body joints and global XYZ root translation. Following prior work~\cite{motron, ma2025uncertainty}, we downsampled sequences by a factor of two to 25~Hz to reduce redundancy.

Following the benchmark subject splits~\cite{motron}, we reserved S9/S11 for probabilistic and stochastic evaluation and S5 for deterministic evaluation, training on the remaining subjects. Splitting subjects before constructing windows prevents overlapping frames from entering both training and evaluation. Motron~\cite{motron} selects checkpoints on sparse evaluation-subject windows, and our hyperparameter sweeps (Section~\ref{sec:hyper}), representation ablation (Fig.~\ref{fig:pose}), and deterministic model followed this protocol. Because this lets evaluation data influence which checkpoint is reported, we fixed the hyperparameters and retrained the probabilistic/stochastic model using only training-subject validation data, namely the final 10\% of each training sequence, separated from the fitting frames by a gap of one full input--output window ($H+F=62$ frames for the horizons in Table~\ref{tab:impl}). S9/S11 were thus used only for its final scoring, although its hyperparameters came from the earlier sweeps.

For probabilistic and stochastic evaluation, 150 training sequences yielded 158,179 fitting windows from all valid starts and 687 validation windows (start stride 17). The tails and gaps withheld 14.2\% of training frames from fitting. From the 59 S9/S11 sequences, every 25th window start was kept for evaluation (2601 windows) to reduce redundant scoring of overlapping windows. Strides apply only to window starts, and poses within each window remain at 25~Hz.

For deterministic evaluation, 180 training sequences yielded 200,987 windows. Checkpoint selection used one in every 100 candidate S5 windows (485 windows). These windows were excluded from gradient updates and normalization, but because they guided checkpoint selection, the deterministic scores may be slightly optimistic. Final evaluation randomly sampled 3840 S5 windows at 128 per sequence (256 per action).

\subsection{Implementation}

We first removed the global root translation, which is known to be corrupted~\cite{motron}, as well as static joints exhibiting zero variance across the dataset, leaving 20 joints. We then converted joint angles to the desired pose representation. 

A public repository\footnote{\url{https://github.com/jingerchong/h36m-tools}} replicates the full preprocessing pipeline and provides utilities for FK, 3D skeleton visualization, and metric computation. It also recreates, from the official raw data, the now-unavailable exponential-map archive of H3.6M~\cite{martinez2017human} used by prior work~\cite{ma2025uncertainty, motron}, with a script that verifies numerical consistency.

We generated training pairs using sliding windows of sizes $H$ and $F$. Dimensions with zero variance were then removed. Unlike the initial filtering step, which eliminated entire static joints, this step prunes individual joint dimensions that do not vary (e.g., Y Euler angle of the elbow), while retaining other active dimensions of the same joint. The remaining data was normalized across joint dimensions and time steps to zero mean and unit variance, using statistics computed only from fitting windows, which were also used to initialize the inducing points.

Following Motron~\cite{motron}, we set $H$ and $F$ according to the values in Table~\ref{tab:impl}. Inducing points were initialized from mini-batch K-means cluster centers using \texttt{SciPy} and GP models were implemented with \texttt{GPyTorch}. Training was parallelized via a \texttt{ModelList}, and the total evidence lower bound (ELBO) was optimized using natural gradient descent (NGD) for variational parameters and Adam for hyperparameters. Additionally, cosine learning rate decay was applied to Adam after a warmup period, and early stopping was employed. Table~\ref{tab:impl} summarizes the selected model and training settings. The probabilistic/stochastic model reached the 50-epoch limit, and its training-subject validation loss selected epoch 45. For the deterministic model, early stopping ended training at epoch 34, with the best checkpoint at epoch 27. Code and trained weights are available at \url{https://github.com/jingerchong/struct-gp}.

\begin{table}[h]
\caption{GP implementation parameters for Probabilistic/Stochastic and Deterministic evaluations. Input and output horizons are reported as number of frames, with the equivalent duration in seconds in parentheses. Horizons follow the Motron benchmark protocols~\cite{motron}, and both configurations use the same compact model size ($M=25$, $L=3$).}
\centering
\setlength{\tabcolsep}{3.5pt}
\begin{tabular}{lcc}
\toprule
Parameter & Probabilistic/Stochastic & Deterministic \\
\midrule
Input $H$ & 12 (0.5~s) & 50 (2~s) \\
Output $F$ & 50 (2~s) & 25 (1~s) \\
Inducing $M$ & 25 & 25 \\
Latent $L$ & 3 & 3 \\
\midrule
Validation data & Train-subject tails & S5 subset \\
Batch (K-means/train/val) & \multicolumn{2}{c}{4096/2048/512} \\
\midrule
NGD LR & 0.4 & 0.45 \\
Adam LR max  & 0.04 & 0.075 \\
Warmup/patience/max epochs & \multicolumn{2}{c}{3/7/50} \\
Early-stop min delta & \multicolumn{2}{c}{0.0005} \\
\midrule
Seeds & \multicolumn{2}{c}{0 (sweeps: 0--2)} \\
GPU & \multicolumn{2}{c}{RTX 3070 Ti} \\
\bottomrule
\end{tabular}
\label{tab:impl}
\end{table}

\subsection{Metrics}

We evaluated predictions with probabilistic, deterministic, and stochastic metrics for comparison with prior work~\cite{motron, ma2025uncertainty}. Baseline results are as reported in the cited works, which use the same subject splits and prediction horizons. Model efficiency was assessed through parameter count and inference cost. Per-window scores are averaged over all evaluation windows.

\textbf{Probabilistic.} Following the evaluation procedure of Motron~\cite{motron}, we generated 2000 samples of predicted rotations, converted them to 3D joint positions via FK, and evaluated 16 joints. KDE NLL was computed in batches of 20 windows, clipped at 20~nats per joint, and summed across joints at each prediction step. The continuous ranked probability score (CRPS) used the first 50 samples. Empirical coverage at the 50\%, 80\%, and 95\% levels used 1000 samples and is the fraction of ground-truth joint coordinates within the corresponding central quantile intervals.

\textbf{Deterministic.} To quantify rotational accuracy, the predicted rotation was computed as the mean rotation in $\text{SO}(3)$ over 50 samples. This mean rotation was then converted to intrinsic ZYX Euler angles in radians, and the L2 norm of the error was obtained over the flattened vector spanning all joints and dimensions, reported as mean angle error (MAE) in radians. Averaging in $\text{SO}(3)$ ensures that the mean rotation respects the geometry of the rotation manifold rather than simply averaging Euler angles in Euclidean space.

\textbf{Stochastic.} We reported best-of-50 average displacement error (ADE) and final displacement error (FDE), as well as average pairwise distance (APD) computed over all 50 samples, in the same 3D joint space as the probabilistic evaluation with the same joints ignored.

\textbf{Model Efficiency.} We counted the number of learnable parameters as a measure of model complexity and memory footprint. Runtime was measured separately for the analytic-moment and sampling stages at batch size one, as described in Section~\ref{sec:efficiency}.

\subsection{Hyperparameter Selection}
\label{sec:hyper}

Table~\ref{tab:ablation} lists the settings explored, with learning rates tuned for each. A three-seed sweep (seeds 0--2) on a 1~s horizon ($H=12$, $F=25$) was evaluated with both the probabilistic and deterministic metrics, and single-seed sweeps at the final horizons in Table~\ref{tab:impl} confirmed its trends for both configurations. In the three-seed sweep, which compared the kernel, $M$, and $L$ using rotations relative to the last observed frame, the standard deviation across seeds was at most 0.17~nats in horizon-averaged KDE NLL (0.62~nats for absolute exponential maps) and 0.005~rad in MAE. Against this variation, absolute 6D rotations outperformed absolute exponential maps and quaternions by 10--11~nats per step and also achieved the lowest MAE among absolute representations. Kernel choice changed KDE NLL by at most 0.03~nats, and increasing $M$ from 25 to 200 gained only 0.42~nats, which motivated the compact $M=25$. A single latent function was 1.28~nats worse than $L=2$, while larger $L$ gained at most 0.17~nats, which motivated $L=3$. Section~\ref{sec:coupling} reports the pose-representation and joint-coupling ablations at the final horizon.

\begin{table}[h]
\caption{Explored settings for representation and hyperparameter selection. The final configuration (6D rotations, Matérn 3/2 + Linear, $M=25$, $L=3$) combines the best-performing representation with a compact model size (Section~\ref{sec:hyper}).}
\centering
\setlength{\tabcolsep}{3.5pt}
\begin{tabular}{lc}
\toprule
Factor & Options Tested  \\
\midrule
Pose representation & exponential map, quaternion, 6D rotations  \\
Covariance kernel & RBF, Matérn 3/2, Matérn 5/2 (+ Linear)  \\
Inducing $M$ & 10, 25, 50, 100, 200  \\
Latent $L$ & 1, 2, 3, 4, 6  \\
Batch size & 256, 512, 1024, 2048, 4096\\
\bottomrule
\end{tabular}
\label{tab:ablation}
\end{table}

\section{RESULTS}

\subsection{Probabilistic Evaluation}
\label{sec:prob}

\begin{figure}[thpb]
  \centering
  \setlength{\fboxrule}{0pt}
  \fbox{\includegraphics[width=.975\linewidth]{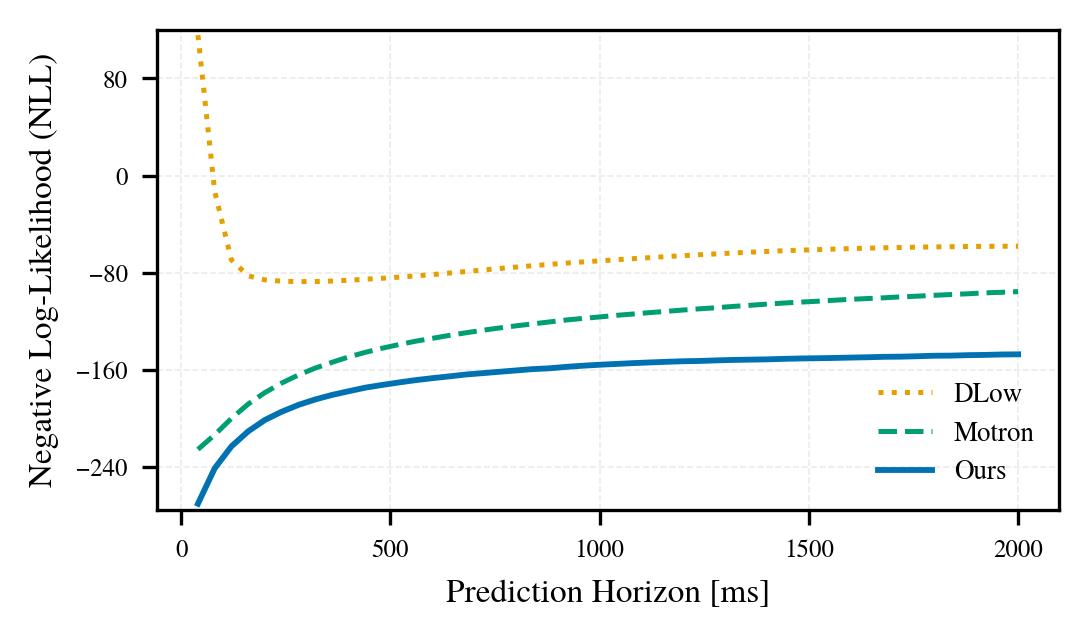}}
  \caption{KDE NLL (lower is better) of our final model, Motron, and DLow. Our model is lowest at every time step and 22--52~nats below Motron, placing substantially higher density on the observed future motion across the full 2~s horizon.}
  \label{fig:final}
\end{figure}

As shown in Fig.~\ref{fig:final}, our method achieves lower KDE NLL than DLow~\cite{dlow} and Motron~\cite{motron} at every prediction step. At early time steps, DLow assigns low density to the ground truth (positive NLL), whereas our model assigns high density (negative NLL). Across the 2~s horizon, the joint-summed KDE NLL of the proposed model is 22--52~nats lower than that of Motron. Over the 16 evaluated joints, this corresponds to an average of 1.4--3.2~nats per joint. The same configuration trained and selected under the original protocol was 5.7--8.0~nats worse at every step, so the margin over the baselines does not depend on selecting checkpoints on the evaluation subjects.

\begin{figure}[thpb]
  \centering
  \setlength{\fboxrule}{0pt}
  \fbox{\includegraphics[width=.975\linewidth]{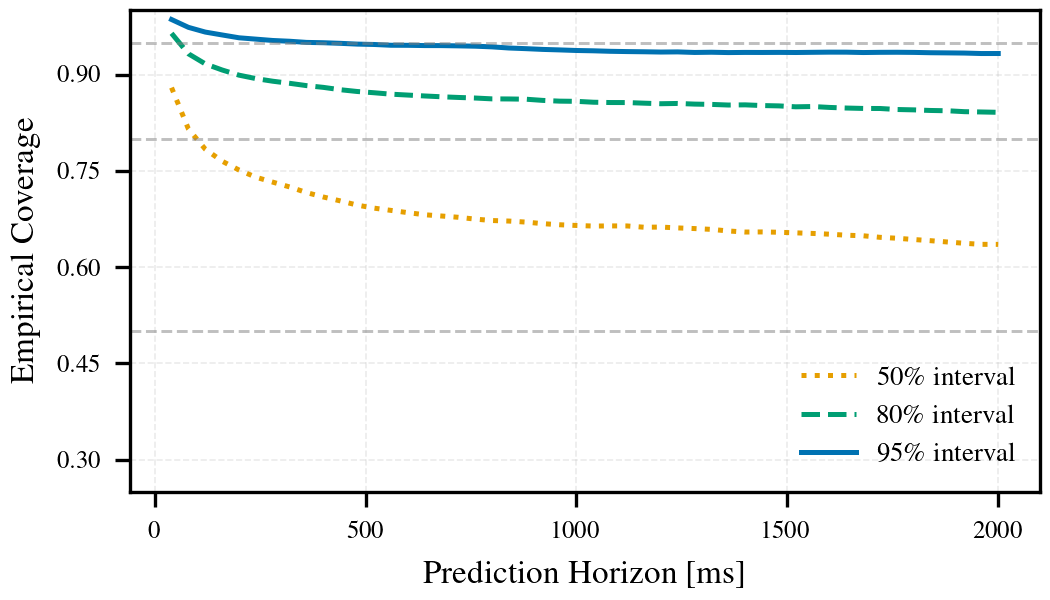}}
  \caption{Empirical coverage of predicted joint positions for the 50\%, 80\%, and 95\% central intervals across prediction horizons. Gray dashed lines mark the nominal levels. The 50\% and 80\% intervals stay above nominal throughout, and the 95\% interval stays near nominal, falling slightly below it (to about 93\%) beyond about 0.5~s.}
  \label{fig:coverage}
\end{figure}

Fig.~\ref{fig:coverage} shows empirical coverage in position space. The 50\% and 80\% intervals over-cover at all horizons, most strongly for the 50\% interval at short horizons, while the 95\% interval stays close to nominal, above it up to about 0.5~s and slightly below it thereafter, reaching about 93\% at 2~s. Coverage changes only gradually with horizon. The model therefore errs on the conservative side at the 50\% and 80\% levels, producing wider regions than nominal. Together with the KDE NLL results, these findings show that the model delivers high-likelihood predictive distributions with conservative-to-near-nominal coverage.

\begin{figure}[thpb]
  \centering
  \includegraphics[width=.975\linewidth]{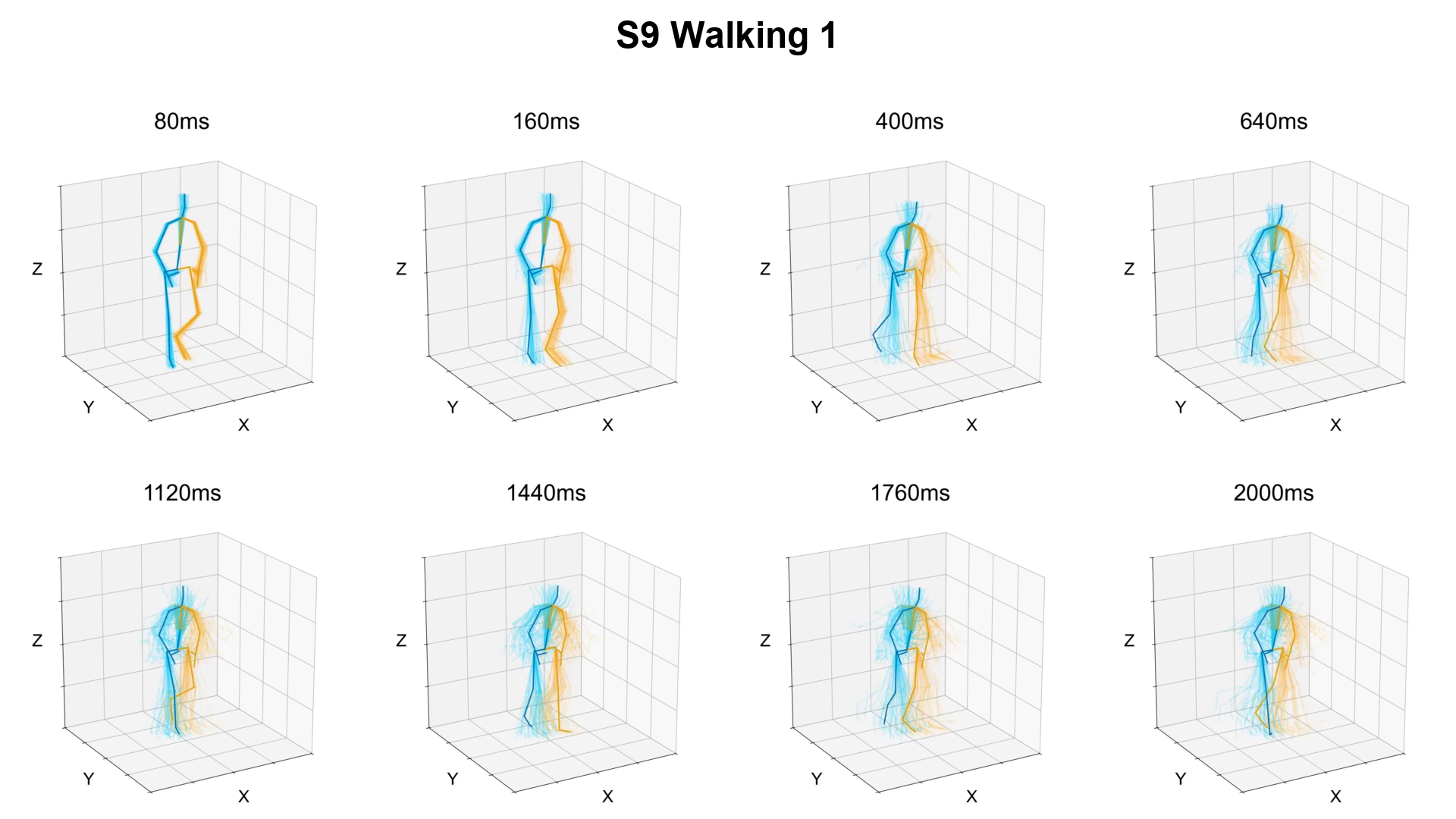}
  \caption{Visualization of 50 sampled skeleton predictions drawn from the predicted distributions at different horizons. The solid skeleton denotes the ground-truth motion, while the translucent skeletons represent sampled predictions. Example shown for Subject S9 performing the walking action (first sequence). Colors indicate body sides (blue: right, orange: left). The sample spread widens with horizon while continuing to enclose the ground truth, illustrating uncertainty that grows as future motion becomes less predictable.}
  \label{fig:samples}
\end{figure}

To illustrate the predicted uncertainty distributions, Fig.~\ref{fig:samples} shows sampled skeleton predictions from the model at different prediction horizons. The increasing dispersion of the samples reflects the model's growing uncertainty in future motion as the horizon extends. An accompanying video\footnote{\url{https://youtu.be/9IB6VJWT0Nc}} demonstrates these predictions for example motion sequences and across multiple action types.

\subsection{Deterministic Evaluation}
\label{sec:det}

Table~\ref{tab:mae} reports the deterministic performance of our model in terms of MAE, computed on the mean rotations of the predictive distribution. Our MAE is lower than that of the zero-velocity and GRU sup baselines at all reported horizons. Relative to Motron, the strongest baseline, the gap narrows from 22\% at 160~ms to 3\% at 1000~ms. The model also outperforms QuaterNet at 80~ms and stays within 0.02 of it up to 400~ms. The mean of the predictive distribution therefore provides a competitive point estimate at longer horizons, even though the model is trained for distribution quality.

\begin{table}[h]
\caption{MAE (rad, lower is better) of our model compared to SOTA. The gap to Motron narrows to 3\% at 1000~ms, so the predictive mean remains competitive at longer horizons. GRU sup and QuaterNet report results only up to 400~ms.}
\centering
\setlength{\tabcolsep}{3.5pt}
\begin{tabular}{lcccccccc}
\toprule
ms& 80 & 160 & 320 & 400 & 560 & 720 & 880 & 1000 \\
\midrule
Zero Vel~\cite{motron} & 0.40 & 0.70 & 1.11 & 1.25 & 1.46 & 1.63 & 1.76 & 1.84 \\
GRU sup~\cite{martinez2017human} & 0.43 & 0.74 & 1.15 & 1.30 & - & - & - & -\\
QuaterNet~\cite{pavllo2020modeling} & 0.37 & 0.62 & 1.00 & 1.14 & - & - & - & - \\
HisRepItself~\cite{mao2020history} & \textbf{0.28} & 0.52 & 0.88 & 1.02 & 1.23 & \textbf{1.40} & 1.55 & 1.64 \\
Motron~\cite{motron} & \textbf{0.28} & \textbf{0.51} & \textbf{0.87} & \textbf{1.01} & \textbf{1.22} & \textbf{1.40} & \textbf{1.54} & \textbf{1.63} \\
\midrule
Ours & 0.33 & 0.62 & 1.02 & 1.16 & 1.36 & 1.50 & 1.61 & 1.68 \\
\bottomrule
\end{tabular}
\label{tab:mae}
\end{table}

\begin{figure}[thpb]
  \centering
  \includegraphics[width=.975\linewidth]{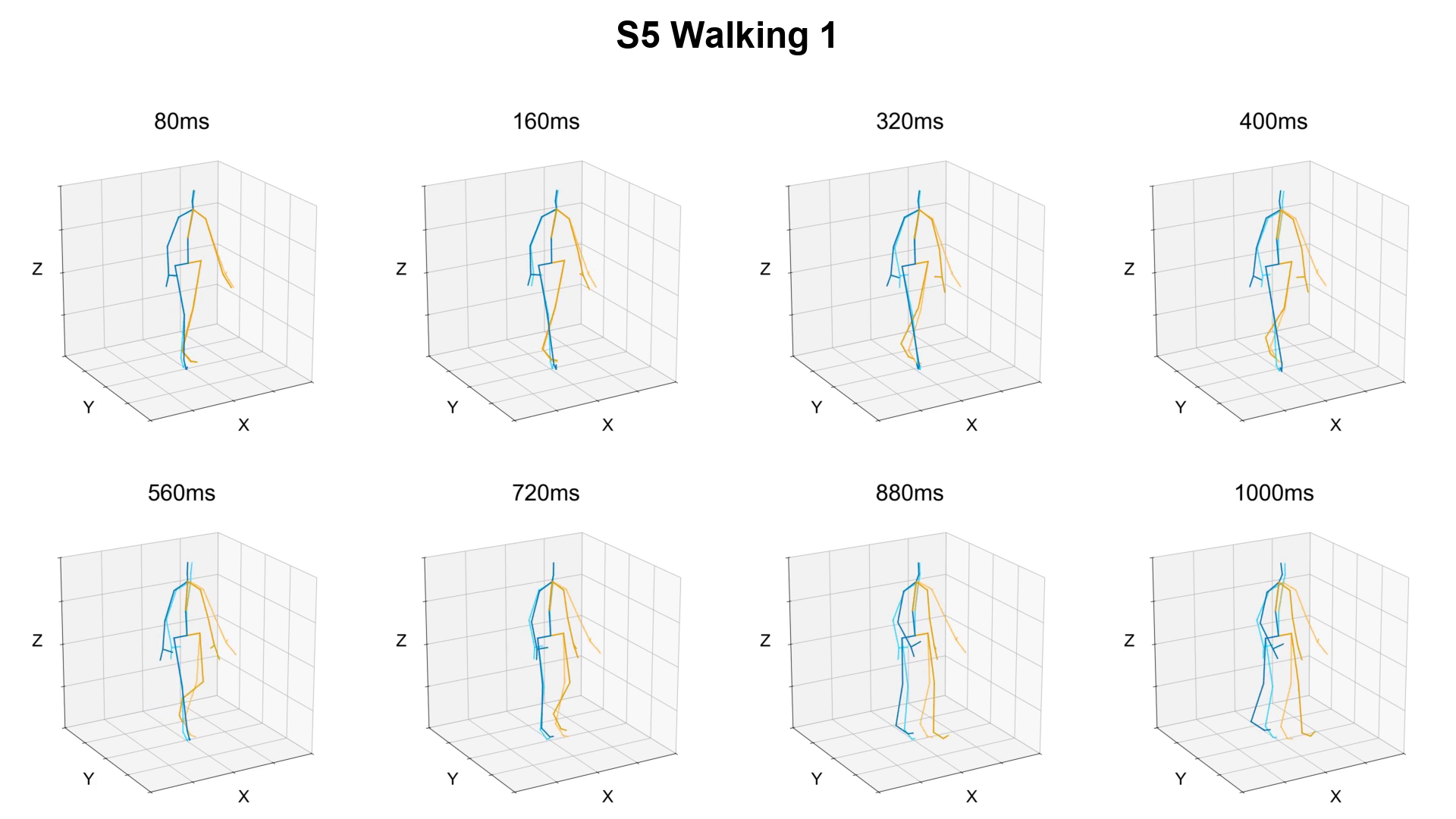}
  \caption{Visualization of mean skeleton predictions at different horizons. The solid skeleton denotes the ground-truth motion, while the translucent skeleton represents the mean prediction. Example shown for Subject S5 performing the walking action (first sequence). Colors indicate body sides (blue: right, orange: left). The mean prediction closely follows the ground truth at short horizons, with deviations growing over time, consistent with the MAE trend in Table~\ref{tab:mae}.}
  \label{fig:mean_skeletons}
\end{figure}

Fig.~\ref{fig:mean_skeletons} highlights the evolution of the mean predicted poses over different horizons. Consistent with the increasing MAE reported in Table~\ref{tab:mae}, the predictions gradually deviate from the ground-truth motion at longer horizons. The accompanying video also shows these predictions over short motion sequences and across action types.

\subsection{Stochastic Evaluation}

Table~\ref{tab:stoch} presents trajectory-level metrics ADE, FDE, and APD, computed over 50 sampled trajectories. Each joint--dimension GP is sampled jointly across its full prediction horizon, preserving its modeled temporal covariance. Cross-joint dependencies are factorized, so joints vary independently within each sampled pose. The best-of-50 ADE and FDE of the proposed model are 13--58\% higher than those of the best baseline at each horizon, and its APD is the lowest at 1~s and second-lowest at 2~s, indicating less diverse samples. These metrics reward diverse sample sets, whereas the model is trained for predictive likelihood, on which it outperforms Motron and DLow (Section~\ref{sec:prob}).

\begin{table}[h]
\caption{APD, best-of-50 ADE, and best-of-50 FDE (m) of our model compared to SOTA. Our samples are less diverse than most baselines, with higher best-of-50 ADE than all baselines and higher FDE than most. These metrics reward diverse sample sets, whereas our model is trained for predictive likelihood, on which it outperforms Motron and DLow (Fig.~\ref{fig:final}).}
\centering
\setlength{\tabcolsep}{3.5pt}
\begin{tabular}{lcccccc}
\toprule
& \multicolumn{3}{c}{1~s} & \multicolumn{3}{c}{2~s} \\
\cmidrule(lr){2-4} \cmidrule(lr){5-7}
& APD $\uparrow$ & ADE $\downarrow$ & FDE $\downarrow$ & 
APD $\uparrow$ & ADE $\downarrow$ & FDE $\downarrow$  \\
\midrule
DSF~\cite{yuan2019diverse} & - & - & - & 9.330 & 0.493 & 0.592 \\
MT-VAE~\cite{mtvae} & - & - & - & 0.403 & 0.457 & 0.595 \\
MotionDiff~\cite{motiondiff} & - & - & - & \textbf{15.353} & 0.411 & 0.509 \\
HumanMAC~\cite{chen2023humanmac} & - & - & - & 6.301 & 0.369 & \textbf{0.480} \\
ProbHMI~\cite{ma2025uncertainty} & - & - & - & 6.682 & \textbf{0.364} & 0.493 \\
DLow~\cite{dlow} & \textbf{5.180} & 0.305 & 0.419 & 11.741 & 0.425 & 0.518 \\
Motron~\cite{motron} & 3.453 & \textbf{0.252} & \textbf{0.350} & 7.168 & 0.375 & 0.488\\
\midrule
Ours & 3.146 & 0.398 & 0.446 & 5.703 & 0.544 & 0.541  \\
\bottomrule
\end{tabular}
\label{tab:stoch}
\end{table}

\subsection{Model Efficiency}
\label{sec:efficiency}

Table~\ref{tab:param} compares model sizes. Our probabilistic/stochastic and deterministic variants use 0.24~M and 0.35~M parameters, respectively, reflecting their different input and output dimensions. Both are comparable in size to siMLPe's deterministic DCT+MLP~\cite{guo2023back} and ProbHMI's autoregressive GCN/RNN~\cite{ma2025uncertainty}. The probabilistic/stochastic variant uses roughly seven times fewer parameters than Motron~\cite{motron} (1.67~M) while achieving lower KDE NLL (Section~\ref{sec:prob}), and the deterministic variant's MAE is 3--22\% higher than Motron's with about five times fewer parameters (Section~\ref{sec:det}).

\begin{table}[h]
\caption{Parameter counts of our model and SOTA methods. Bold marks the fewest parameters among probabilistic and stochastic methods. Our probabilistic model is about seven times smaller than Motron while achieving lower KDE NLL, showing that high-quality predictive distributions can be obtained with a compact parameterization.}
\centering
\setlength{\tabcolsep}{3.5pt}
\begin{tabular}{lccc}
\toprule
& Type & Method & Params (M) \\
\midrule
siMLPe~\cite{guo2023back} & Deterministic & DCT, MLP & 0.14 \\
HisRepItself~\cite{mao2020history} & Deterministic & Attention, GCN & 3.24 \\
DLow~\cite{dlow} & Stochastic & Latent, CVAE & 7.30 \\
MotionDiff~\cite{motiondiff} & Stochastic & Diffusion & 29.93 \\
HumanMAC~\cite{chen2023humanmac} & Stochastic & DCT, Diffusion & 28.40 \\
Motron~\cite{motron} & Probabilistic & Latent, GNN & 1.67 \\
ProbHMI~\cite{ma2025uncertainty} & Probabilistic & GCN and RNN & 0.31 \\
\midrule
Ours & Prob./Stoch. & GP & \textbf{0.24} \\
& Deterministic & GP & 0.35 \\
\bottomrule
\end{tabular}
\label{tab:param}
\end{table}

Both configurations in Table~\ref{tab:impl} were timed on an RTX 3070 Ti at batch size one, with two warmup passes and ten measured passes. For the probabilistic/stochastic model, computing the analytic means and covariances of all 96 GPs took 997~ms per sequence, and drawing 2000 samples followed by postprocessing and FK took a further 468~ms (1465~ms total). For the deterministic model, the same two stages took 1140~ms and 454~ms (1594~ms total). Per-GP cost was nearly constant from 1 to 96 GPs (9.9--10.7~ms and 11.6--13.5~ms for the two models), so cost scales linearly with the number of GPs in the current sequential \texttt{ModelList} evaluation. Because the GPs are independent, batched or parallel evaluation could reduce latency. Consumers that use the rotation-space Gaussian directly skip the sampling and FK stage. These timings characterize the analytic and sampling stages of the proposed model.

\subsection{Ablation Studies}
\label{sec:coupling}

\begin{figure}[thpb]
  \centering
  \setlength{\fboxrule}{0pt}
  \fbox{\includegraphics[width=.975\linewidth]{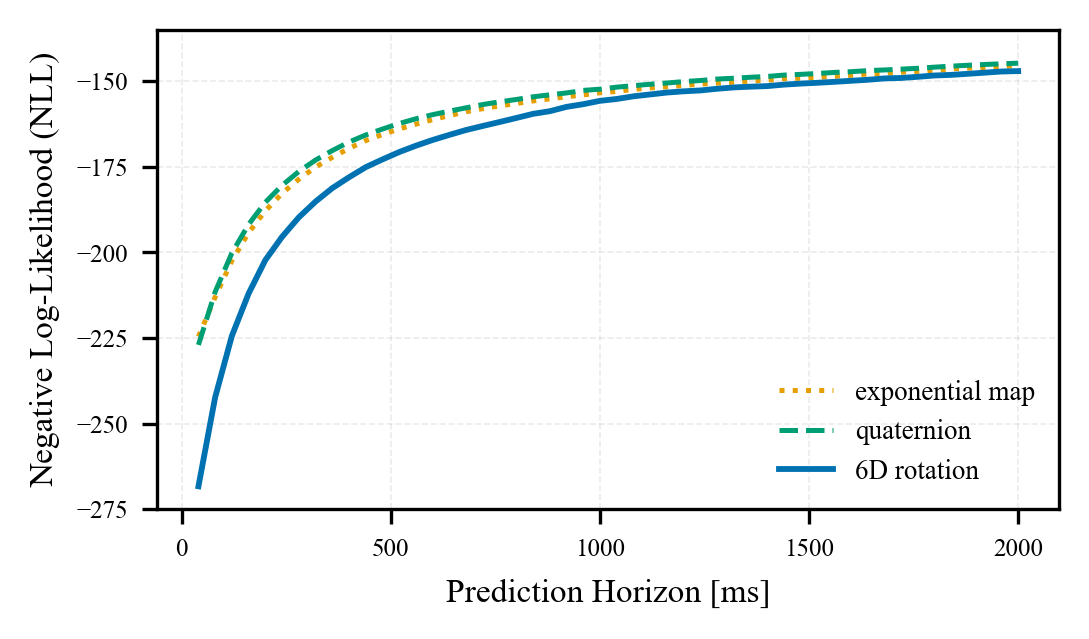}}
  \caption{KDE NLL by horizon for seed-0 absolute-rotation models with $H=12$, $F=50$, a Matérn 3/2 kernel, and $M=100$. Curves use the reported adaptive estimator with 2000 posterior samples. The 6D representation has the lowest NLL at every horizon, supporting its selection for the final model. Its advantage narrows at longer horizons as the adaptive KDE bandwidth grows with predictive spread.}
  \label{fig:pose}
\end{figure}

Fig.~\ref{fig:pose} compares rotation representations at the final horizon. The 6D representation gives the lowest KDE NLL at every step, while exponential maps and quaternions perform similarly. The gap narrows at longer horizons partly because KDE NLL smooths predicted samples with an adaptive bandwidth that grows with predictive spread, and fixing the bandwidth at its first-horizon value changes the ranking. Growing predictive variance also reduces the influence of local smoothness differences between rotation parameterizations. A metric that requires no density smoothing confirms the advantage of 6D: its 2~s best-of-50 ADE is 0.537~m versus 0.641~m and 0.652~m for exponential maps and quaternions.

To test joint--dimension factorization, we compared 14 independent GPs (A), three within-joint GPs (B), and one full-chain GP (C) for the right shoulder, elbow, and wrist, retaining 14 rotation coordinates. All used the horizons in Table~\ref{tab:impl} but, unlike the primary model, an RBF kernel, $M=200$, and a latent/task ratio of 0.5, trained for 50 epochs of ten batches of 512 sequences. Table~\ref{tab:coupling} reports scores on 40 test windows, with non-arm rotations supplied from the target during FK. KDE NLL and CRPS cover three arm endpoints. Parameter counts are dominated by the $M=200$ variational covariances and are therefore far larger than those of the main model. Coupling increased NLL by 6.42 and 8.62 and CRPS by 0.0426 and 0.0614~m for B and C relative to A, although coverage was closer to nominal. Coupling also raised peak GPU memory from 1.52~GiB (A) to 3.15~GiB (C) and, for C, the epoch time from 5.59~s to 20.24~s. At this fixed training budget, factorization therefore delivered the best predictive scores at the lowest memory cost, supporting joint--dimension independence as an effective and efficient design.

\begin{table}[h]
\caption{Right-arm coupling at a shared training budget for independent (A), within-joint (B), and full-chain (C) GPs. Memory is peak allocated GPU memory, and epoch time covers ten batches. The factorized variant A achieves the best KDE NLL, CRPS, and MAE with the lowest memory, supporting joint--dimension factorization in the main model.}
\centering
\setlength{\tabcolsep}{3.5pt}
\begin{tabular}{lccc}
\toprule
& A & B & C \\
\midrule
KDE NLL & -11.237 & -4.814 & -2.613 \\
CRPS (m) & 0.0473 & 0.0899 & 0.1087 \\
MAE (rad) & 0.637 & 0.776 & 1.021 \\
Coverage 50\% & 70.6\% & 58.6\% & 49.0\% \\
Coverage 80\% & 91.0\% & 84.2\% & 79.4\% \\
Coverage 95\% & 97.1\% & 96.0\% & 95.4\% \\
Parameters (M) & 14.127 & 14.223 & 14.409 \\
Epoch time (s) & 5.59 & 5.29 & 20.24 \\
GPU memory (GiB) & 1.52 & 2.08 & 3.15 \\
\bottomrule
\end{tabular}
\label{tab:coupling}
\end{table}

\section{CONCLUSION}

This paper introduced a structured multitask variational GP that brings closed-form, coverage-checked uncertainty to full-body human motion prediction. By combining one-shot horizon prediction, multitask temporal coupling, joint--dimension factorization, and 6D rotations, the model yields Gaussian trajectories with full temporal covariance. Planners operating in rotation coordinates can consume these distributions directly, without density estimation or the 468~ms sampling and FK stage. On H3.6M, the predicted distributions assign higher density to the ground truth than Motron and DLow throughout the 2~s horizon, with 22--52~nats lower KDE NLL than Motron and coverage that is conservative at 50\% and 80\% and near-nominal at 95\%. The probabilistic model uses only 0.24~M parameters, and a separately trained 0.35~M-parameter deterministic variant has an angle error 3--22\% higher than Motron's, a gap that narrows to 3\% at 1~s. The ablations support the 6D representation and joint--dimension factorization.

The evaluation covers a single benchmark of single-subject activities, so benefits for robot behavior remain to be shown. Future work will integrate the distributions into closed-loop planners and compare probabilistic and deterministic predictions on interaction-conditioned human--robot tasks. Extending the formulation beyond the fixed H3.6M skeleton to other datasets and articulated embodiments requires redefining the joint--dimension set and FK tree. Further directions include full-body cross-joint dependencies, longer horizons, robustness to missing observations, and parallel GP evaluation, timed against deep baselines under a common protocol. Overall, structured GPs offer a compact and analytically tractable route to human motion uncertainty with conservative empirical coverage for robotic systems.

\addtolength{\textheight}{-12cm}   

\bibliographystyle{IEEEtran}
\bibliography{IEEEabrv, ref}

\end{document}